\documentclass[review,11pt]{ReportTemplate}
\usepackage{bm}
\usepackage{graphicx}
\usepackage{paralist, algorithmic, algorithm}
\usepackage{amsmath, mathrsfs, amssymb}
\usepackage{subfigure}
\usepackage{geometry}
\usepackage{amsmath, amssymb}
\usepackage{multirow}
\usepackage{arydshln}  
\usepackage{longtable, booktabs}
\usepackage{float}
\usepackage[permil]{overpic}
\usepackage{color, xcolor}
\usepackage{helvet}   
\usepackage{setspace}
\usepackage{natbib}   
\usepackage{caption}  
\usepackage{newfloat}
\usepackage{listings}

\definecolor{steelblue}{RGB}{46, 82, 180}

\begin{document}
\begin{frontmatter}

\title{Rethinking the Value of Labels for Instance-Dependent Label Noise Learning}

% The \author macro works with any number of authors. There are two commands
% used to separate the names and addresses of multiple authors: \And and \AND.
%
% Using \And between authors leaves it to LaTeX to determine where to break the
% lines. Using \AND forces a line break at that point. So, if LaTeX puts 3 of 4
% authors names on the first line, and the last on the second line, try using
% \AND instead of \And before the third author name.

\author{%
  Hanwen Deng$^1$, $\space$ Weijia Zhang$^3$, $\space$ Min-Ling Zhang$^{1}$\corref{cor1}\\
   $^1$School of Computer Science and Engineering, Southeast University, \\Nanjing 210096, China\\
   $^2$Key Laboratory of Computer Network and Information Integration (Southeast University), \\Ministry of Education, China\\
   $^3$ School of Information and Physical Sciences, The University of Newcastle,\\ Callaghan, NSW 2308, Australia\\
  \texttt{$\{$denghw,zhangml$\}$@seu.edu.cn,weijia.zhang@newcastle.edu.au} \\
  % examples of more authors
  % \And
  % Coauthor \\
  % Affiliation \\
  % Address \\
  % \texttt{email} \\
  % \AND
  % Coauthor \\
  % Affiliation \\
  % Address \\
  % \texttt{email} \\
  % \And
  % Coauthor \\
  % Affiliation \\
  % Address \\
  % \texttt{email} \\
  % \And
  % Coauthor \\
  % Affiliation \\
  % Address \\
  % \texttt{email} \\
}
\cortext[cor1]{ Corresponding author.}

\begin{abstract}
  Label noise widely exists in large-scale datasets and significantly degenerates the performances of deep learning algorithms.
   Due to the non-identifiability of the instance-dependent noise transition matrix, most existing algorithms address the problem by assuming the noisy label generation process to be independent of the instance features.
   Unfortunately, noisy labels in real-world applications often depend on both the true label and the features.
   In this work, we tackle instance-dependent label noise with a novel deep generative model that avoids explicitly modeling the noise transition matrix.
   Our algorithm leverages casual representation learning and simultaneously identifies the high-level content and style latent factors from the data.
   By exploiting the supervision information of noisy labels with structural causal models, our empirical evaluations on a wide range of synthetic and real-world instance-dependent label noise datasets demonstrate that the proposed algorithm significantly outperforms the state-of-the-art counterparts.
\end{abstract}
\end{frontmatter}

\section{Introduction}

Deep neural networks have achieved remarkable performances in many tasks thanks to the availability of large datasets with high-quality labels.
However, the time and monetary costs of accurately annotating large datasets are often prohibitively high in applications such as medical analysis ~\cite{karimi:deep} and social science ~\cite{cothey:web}. 
To alleviate this issue, crowd-sourced labels are frequently utilized which generally provide low-quality inaccurate labels \cite{irvin:chexpert,xiao:learning,zhou:open}.
Even for tasks where the labeling cost is relatively low, it is often unavoidable that label noise will be introduced during the collection of datasets ~\cite{yan:learning}.
Since noisy labels will significantly deteriorate the generalization performance of deep neural network models ~\cite{arpit:closer}, developing an efficient algorithm for learning from noisy labels has gradually received increasing attention from the community ~\cite{xia:anchor,cheng:learning,li:dividemix,han:co,tanaka:joint}.

Most existing work tackles label noise by assuming that the noisy labels are independent of the instance features, i.e., instance-independent noise (IIN).
Furthermore, IIN algorithms either assume random classification noise where the incorrect labels are assigned randomly ~\cite{natarajan:learning,manwani:noise}, or class-conditional noise where each instance has a fixed probability of being assigned to another class depending on its true class label ~\cite{xia:anchor,zhang:generalized}.
Although several algorithms that model the noise transition matrix have been proven to be theoretically consistent; 
however, the assumption that noises are independent of the instances is not practical in the real-world ~\cite{cheng:learning}.
%\note{A note on choosing references when writing paper: consider two perspectives: the historical first/important papers on the topic (to make your paper to read more professional), and the researchers that currently active (to slightly increase your paper's chance for acceptance).}

Instance-dependent noise (IDN) is a more general assumption that is suitable for real-world applications ~\cite{xia:part,cheng:learning,chen:beyond}, where the noisy labels depend not only on the true \emph{class} of the instances but also on its \emph{features}.
IDN occurs naturally when we consider the data annotation process, i.e., the instance label in crowdsourcing is provided by an annotator who outputs a label based on observing its instance feature and the pre-defined set of candidate labels. 
Therefore, the noisy labels clearly depend on the feature. 
For webly-labeled data, it has also been shown that the class-conditional noise assumption is statistically impossible to be valid ~\cite{chen:beyond} in well-known datasets such as Clothing1M ~\cite{xiao:learning}.

IDN poses two significant challenges.
Firstly, algorithms that rely on the small loss criterion are no longer effective because the noisy labels are highly correlated with the features which significantly reduces the memorization effect ~\cite{chen:beyond,jiang:beyond}. 
Secondly, consistent algorithms that identify the noise transition matrix are also not suitable because the high-dimensional feature makes identification an ill-posed problem ~\cite{xia:anchor}. 
Therefore, their performances degenerate significantly in real-world applications with instance-dependent noise ~\cite{chen:beyond}.
An early attempt solves the problem by assuming part-dependent noise ~\cite{xia:part}, which states that instances can be decomposed into parts and the noise transition matrix is defined by parts. 
However, such assumptions are restrictive and difficult to verify ~\cite{cheng:instance}. 
Other attempts avoid explicitly estimating the transition matrix by modeling the causal relationship among the instances and the labels; unfortunately, it requires modeling the causal relationships among the high dimensional features and the labels.

In this paper, we address the IDN problem from the novel perspective of causal representation learning ~\cite{scholkopf:toward}.
Unlike previous methods that primarily work with the raw input features, we instead propose to exploit the valuable weak supervision signal conveyed in the instance-dependent noisy labels which enables us to work with low-dimensional causal representations that separately capture the \emph{content} and \emph{style} semantics of the instances.
The \emph{content} latent captures the causal information corresponding to the unobserved clean label. In contrast, the \emph{style} latent captures characteristics that are not causally related to the true label, but may contribute towards the generation of noisy labels (Figure \ref{fig:Graphical causal model}).
Guided by a structural causal model and implemented through Variational Auto-Encoders (VAEs) \cite{kingma:vae,joy:capturing}, our generative model not only exploits the theoretical and empirical success of weakly supervised causal representation learning ~\cite{von:self,zhang:mil}, but also aligns with psychological and physiological evidence that human annotators do not perceive objects based on the raw input signal from the retina, but instead rely on semantic concepts processed by the visual cortex ~\cite{logothetis:visual,ullman:high}.

The key idea of how we utilize noisy labels significantly differs from both supervised/semi-supervised VAEs and existing IDN approaches in two aspects.
Firstly, unlike previous VAE-based methods that aim to associate certain latent variables with labels \cite{sohn:learning}, our method captures characteristic information conveyed by the noisy supervision, and thus preserves information much richer than only the labels themselves.
Secondly, different from previous IDN methods that use co-teaching to select possibly clean labels \cite{yao:instance}, our approach uses co-teaching to infer the conditional mixture latent priors from the noisy labels, and thus guarantees latent identifiability.

Therefore, we coin our model REthinking Instance-Dependent noisy label Variational AutoEncoder (ReIDVAE).
An implementation of ReIDVAE is provided in the supplementary material and will be made publicly available after publication. 

We summarize our main contributions as follows:
\begin{itemize}
    \item We propose a practical structural causal model on the data generation process of instance-dependent label noise by dividing the latent space into \emph{content} and \emph{style}, and avoid the non-identifiable instance-dependent noise transition matrix problem.
    \item We propose a novel VAE-based deep generative model that avoids encapsulating the labels themselves and ensures capturing the characteristics information of the labels, while simultaneously disentangling the latent space with noisy supervision signals.
 % By dividing the latent variables into content components and style components, the rich characteristic information associated with classification can be captured through labels, thus effectively reducing the impact of noise data.
    \item We experimentally validate that our method achieves superior performances on multiple IDN benchmarks with controlled noise levels including FashionMNIST, SVHN, CIFAR-10, and CIFAR-100 with different levels of instance-dependent noise, and real-world noisy datasets including CIFAR-10N and Clothing1M.
\end{itemize}

\section{RELATED WORK}
\textbf{Random Noise} 
Based on this observation, 
Co-teaching~\cite{han:co} designs two parallel networks that aim to select clean training samples for each other; 
Dividemix~\cite{li:dividemix} fit a two-component Gaussian Mixture Model to distinguish clean and noisy samples;
ELR~\cite{liu:early} designs a regularization term that is robust to noise labels and promotes the learning of clean labeled data.

\noindent \textbf{Class-dependent Noise}
Methods that are designed for addressing class-dependent label noise attempt to estimate the noise transition matrix among different sample classes, i.e., 
$T_{i,j}(\bm{x})=P(\tilde{Y}=j|Y=i)$. 
Given the transition matrix, the clean class posterior probability $P(Y|X=\bm{x})$ can be inferred.
Masking~\cite{han:masking} conveys human cognition of invalid class transitions and speculates the structure of the noise transition matrix.
Forward~\cite{patrini:forward} proposes a framework using two loss corrections to estimate the transition matrix more accurately.
T-Revision~\cite{xia:anchor} first initializing the transition matrix by exploiting examples similar to the anchor points~\cite{yu:learning}, then modifying the transition matrix by a slack variable. 
D2L~\cite{ma:dimensionality} proposes a strategy that identifies the learning epoch at which the transition from dimensional compression to dimensional expansion occurs.

\noindent \textbf{Instance-dependent Noise}
Compared to random and class-dependent noise, instance-dependent noise closely resembles the generation process of real-world noisy labels. 
% These methods attempt to model the instance-dependent label noise. 
Due to the ill-posedness of the instance-dependent noise transition matrix, the modeling processes of most IDN methods are accompanied by strong assumptions. 
For example, BILN~\cite{cheng:learning} assumes that the noise rate is bounded, PTD~\cite{xia:part} assumes that the label noises are part-dependent, ILFC~\cite{berthon:confidence} assumes the confidence-scored instance-dependent noise. 
BLTM~\cite{yang:estimating} propose to directly build the transition between Bayes optimal labels and noisy labels. $Cores^2$~\cite{cheng:learning} continuously filters noisy samples in the dataset by adding regularization terms that encourage the model to make confident predictions, as well as thresholds set according to the feature of each sample. 
Unfortunately, the validity of their assumptions is usually difficult to verify when solving real-world problems. Furthermore, their performances degenerate significantly as the noise rates increase.
Perhaps the method most closely related to ours is CausalNL \cite{yao:instance}, which also models the generation process of IDN with a structural causal model. Specifically, CausalNL assumes that $Y$ is the cause of $\bm{x}$ and uses $P(\bm{x})$ to help address instance-dependent label noise. Our ReIDVAE differentiates from CausalNL in two aspects, firstly, our approach assumes a more reasonable process for generating noise labels, and constructs a more reasonable structural causal model. This contribution also makes the derivation of our loss function more natural. Secondly, our approach pays more attention to the practical meaning of the latent variable space and the impact of clean and noisy labels. This enables our method to distinguish different features of instances and better utilize latent variables to reduce the impact of noisy labels on the model training process. As a result, our method can achieve superior performance even under higher noise rates.

\section{METHOD}

\subsection{Notations}
Let $\mathcal{X}=\mathbb{R}^{m}$ denote the $m$-dimensional instance feature space and let  $\mathcal{Y}=\{0,\cdots,C\}$ denote the multi-class label space,
our data ${\tilde{D}}=\{\bm{x}_{i},\tilde{{Y}}_{i}\}_{i=1}^{N}$ consists of $N$ training samples, where each sample is represented by raw feature $\bm{x}_i \in \mathcal{X}$ and label $\Tilde{Y}_i\in \mathcal{Y}$ which may be contaminated with instance-dependent noise.
Each instance $\bm{x}_{i}$ is also associated with an unobserved clean label $Y_i\in \mathcal{Y}$. The goal of the algorithm is to learn the mapping from input $\bm{x}$ to the clean label $Y$.
When the context is clear, we drop the subscript $i$ to avoid cluttering the notations.
Furthermore, we use $\bm{z}=\{\bm{z}_c,\bm{z}_{s}\}$ to denote the latent content and style factors, where $\bm{z}_c=(z_{1},z_{2},...,z_{C}) \in \mathbb{R}^C $ is the latent content factor and $\bm{z}_{s} \in \mathbb{R}^{d}$ is the latent style factor with $d \ll m$.

\begin{figure}
    \centering
    \includegraphics[width=0.31\textwidth,height=2.3 in]{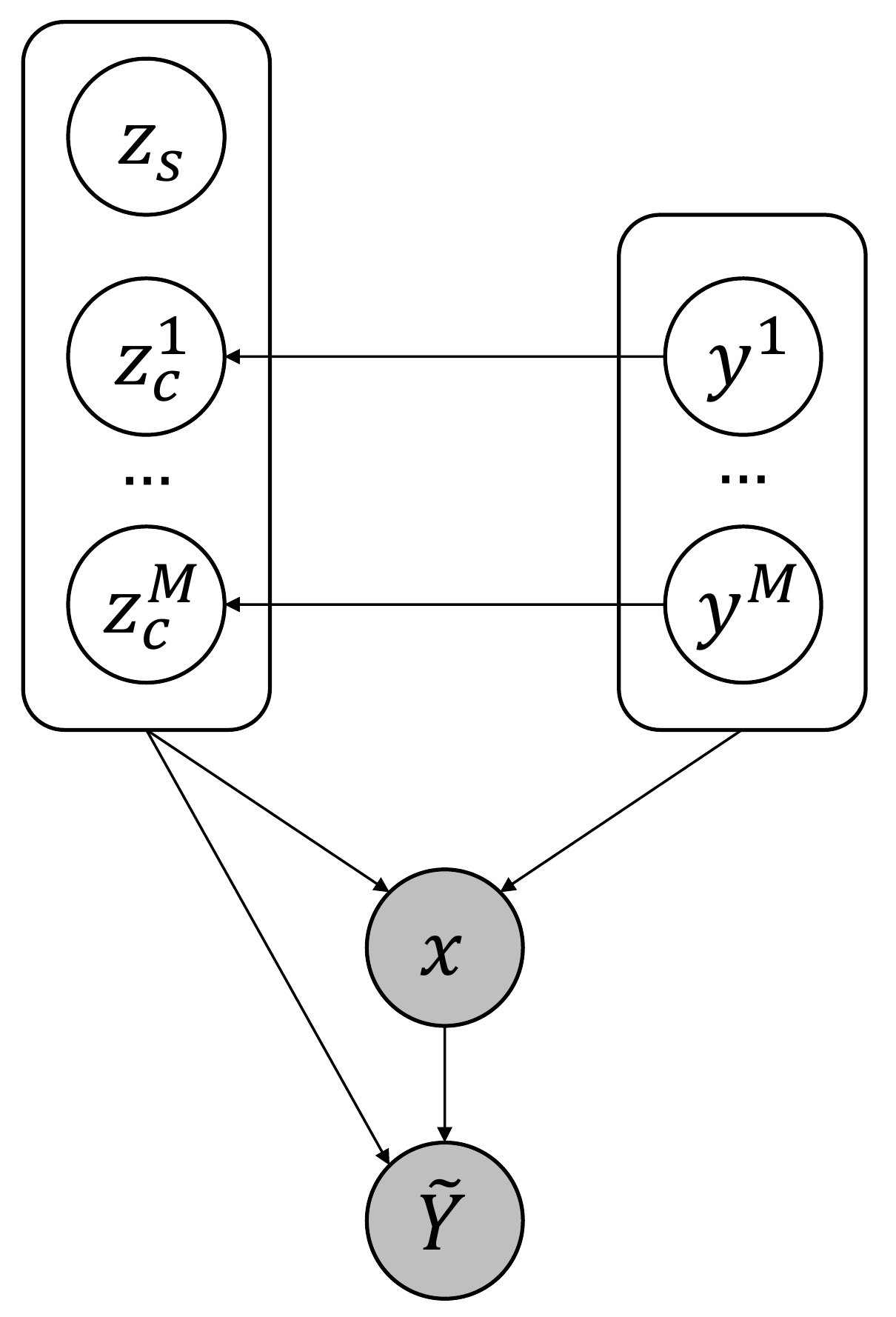}
    \caption{ The generative model of ReIDVAE. The observed information is represented with shaded nodes and latent factors are unshaded. 
    Specifically, $\bm{x}$ represents the raw input feature and $\tilde{Y}$ denotes the noisy label. 
    $Y$ represents the unobservable clean label, 
    and $\bm{z}=\{ \bm{z}_s, \bm{z}_c\}$ corresponds to the latent representation which can be further disentangled into style and content factors.}
     \label{fig:Graphical causal model}
\end{figure}

\subsection{SCM-Inspired Generative Model}
To model the data generation process of datasets with instance-dependent label noise, we assume that the clean label $Y$ is a cause of feature $\bm{x}$ \cite{scholkopf:on}. As illustrated in Figure \ref{fig:Graphical causal model}, the structural causal model (SCM) of the DGP can be written as
%\begin{equation}
%    \begin{split}
%        &Y= \epsilon_Y, \bm{z}=f_{\bm{z}} \left( f_{\bm{z}_c}\left(Y, \epsilon_{\bm{z}_c}\right), \epsilon_{\bm{z}_{s}}\right), \\ 
%        &\bm{x}=f_{\bm{x}} \left( Y, \bm{z},  \epsilon_\bm{x}\right), \tilde{Y}=f_{\tilde{Y}} \left(\bm{x}, \bm{z}, \epsilon_{\tilde{Y}}\right),
%    \end{split}
%\end{equation}
\begin{equation}
    \begin{aligned}
        & Y=\epsilon_Y, \boldsymbol{z}=f_{\boldsymbol{z}}\left(f_{\boldsymbol{z}_c}\left(Y, \epsilon_{\boldsymbol{z}_c}\right), \epsilon_{\boldsymbol{z}_s}\right) \\
        & \boldsymbol{x}=f_{\boldsymbol{x}}\left(Y, \boldsymbol{z}, \epsilon_{\boldsymbol{x}}\right), \tilde{Y}=f_{\tilde{Y}}\left(\boldsymbol{x}, \boldsymbol{z}, \epsilon_{\tilde{Y}}\right)
    \end{aligned}
\end{equation}
where $\epsilon_{\bm{z}_c}$, $\epsilon_{\bm{z}_s}$, $\epsilon_{Y}$, $\epsilon_{\bm{x}}$ and $\epsilon_{\tilde{Y}}$ are independent exogenous variables that determine the error terms \cite{peters:elements};
and the functions $f_{\bm{z}}, f_{\bm{x}}$, and $f_{{\Tilde{Y}}}$ specify the distributions of a variable conditioned on its parents.
On the one hand, from Figure \ref{fig:Graphical causal model} we can see that the clean label ${Y}$ is the cause of the content latent $\bm{z}_c$ but not the style latent $\bm{z}_s$;
on the other hand, the entirety of $\bm{z}$ and ${Y}$ are causes to the observed instance $\bm{x}$. 
Furthermore, $\bm{z}$ and $\bm{x}$ are also causes of the noisy label $\tilde{{Y}}$.

Our SCM is general in the sense that it is suitable for the data generation process of a wide range of datasets. 
For example, during the collection of the MNIST handwritten digits, the writer first determines which number to write; thus dertermining the label ${Y}$.
Then, the writer begins writing according the intrinsic characteristics of the Arabic numeric symbol (content) and the writing style of the writer (style).
Therefore, it is natural to capture the intrinsic characteristics of the pre-determined Arabic number with the content latent factor ${\bm{z}_c}$, and capture the other factors such as writing habits by the style latent factor ${\bm{z}_{s}}$. The above process can be reflected by the structural causal equation of $f_{\bm{z}}$ and $f_{\bm{x}}$.
During the labeling process, the handwritten images are provided to an annotator who outputs a label given an input feature $\bm{x}$. 
However, as supported by physiological and psychological evidence \cite{ullman:high}, human annotators do not predict the instance $\bm{x}$ according to raw pixels. Instead, they decide according to the high-level semantics captured and processed by the visual cortex.
The labeling process is reflected by the structural causal equation $f_{\Tilde{Y}}$.
It is worth noting that $f_{\Tilde{Y}}$ also depends on $\bm{x}$ since the latent factors are designed to capture the high-level semantics instead of all the subtle details of complex objects. Therefore, for instances that are difficult to label, annotators will consider finer details to be a secondary factor \cite{logothetis:visual}.
%It should be noted that in this process, higher-order semantic information plays a dominant role, while the detailed information contained in the original input features plays only an auxiliary role.
%Only for instances that are difficult to label, annotators consider the finer details \cite{logothetis:visual}.

\subsection{Armotized Variational Inference}
Due to the reason that the latent clean label $Y$ is the cause of $\bm{x}$ in our SCM, $P(\bm{x})$ will contain information about $P(Y|\bm{x})$, making $P(\bm{x})$ and $P(Y|\bm{x})$ entangled \cite{scholkopf:causal,zhang:distinguishing}.
To help estimate $P(Y|\bm{x})$ with $P(\bm{x})$, we approximate the SCM using a VAE-based deep generative model. The causal generative process will estimate $P(\bm{x}|Y)$, which benefits for estimating the latent clean label \cite{yao:instance}. 
Under the Markov condition, the joint distribution of the data and latents can be decomposed as:
\begin{equation}
	P(\bm{x},  \tilde{Y}, {Y}, \bm{z})=P({Y})P(\bm{z_c}|{Y}) P_{\theta}(\bm{x},\tilde{Y}|Y,\bm{z}).
\label{Markov}
\end{equation}
Equation \ref{Markov} shows a fundamental difference between our proposed ReIDVAE and existing causality-inspired IDN methods \cite{yao:instance}: our latent $\bm{z}$ depend on a conditional distribution which is crucial for latent identifiability because latents with unconditional isotropic Gaussian prior distributions are proven to be unidentifiable \cite{locatello:challenging}.

\begin{figure*}[!t]
	\centering
	\includegraphics[width=1.0\textwidth]{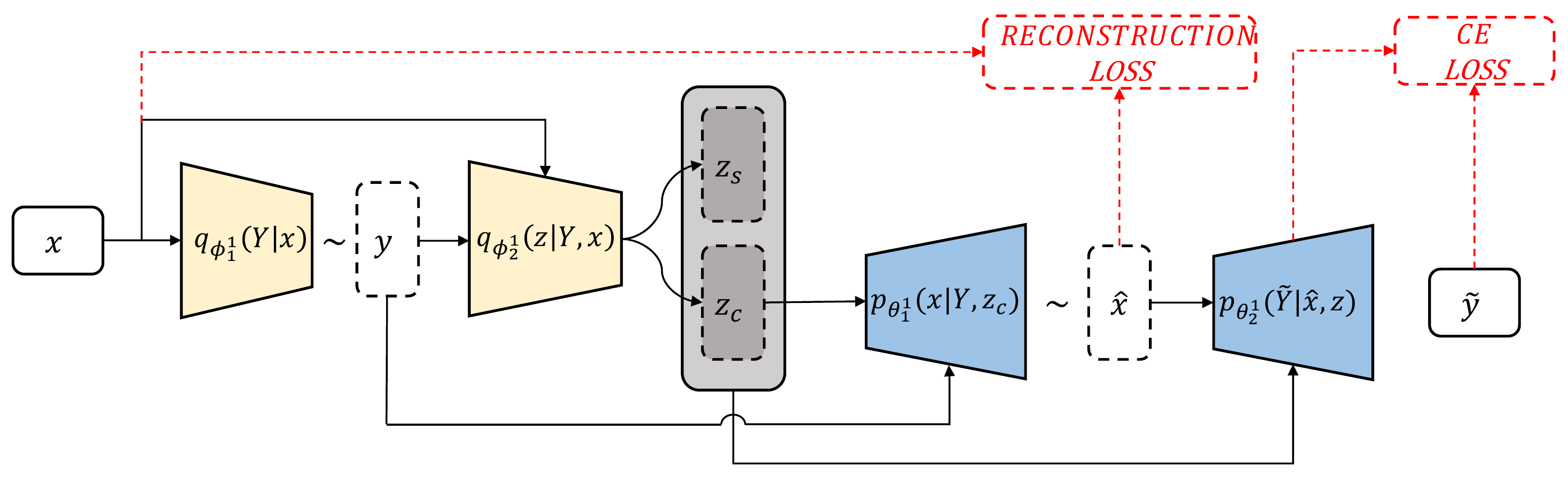}
	\caption{\label{fig:vae model} The structure and workflow of our proposed ReIDVAE method. The yellow and blue shaded modules represent encoders and decoders, respectively. 
    %The red dotted box is part of the loss calculation.
    The first encoder aims to predict the true labels $Y$ of the given instance $\bm{x}$, and then the second encoder uses the predicted $Y$ and the input feature $\bm{x}$ to infer the latent representation $\bm{z}$. 
    Afterward, the style latent $\bm{z}_s$ is excluded from $\bm{z}$ when reconstructing $\hat{\bm{x}}$. 
    Finally, the full latent representation and the reconstructed $\hat{\bm{x}}$ are used to decode the noisy label $\tilde{Y}$. 
    During testing, only $q_{\phi_{1}^1}(Y|\bm{x})$ is needed for predicting unseen instances.
    %The reconstruction loss ensures that we can still reconstruct an instance that is as similar as possible to the original instance even without the style component. The estimation of the noisy label is similar to the noise transition matrix, but since $\tilde{\bm{x}}$ no longer contains noise information, we need to use the complete $\bm{z}$ to estimate $\tilde{Y}$. 
    %\added[id=WZ]{The $x_1$ and $x_2$ for the reconstruction loss should be $\tilde{x}_1$ and $\tilde{x}_2$. We should use solid boxes to represent observed data, e.g., $x$ and $\tilde{y}$; and use grey boxes to represent latent variables, e.g., $z$. Also the vertical location of $z_{\setminus c_1}$, $z_{\setminus c_2}$ is not the same, which is confusing. } 
    }
\end{figure*}

Given the above decomposition of the generative model, we model the marginal of the labels with a categorical distribution. For the conditional latent prior latent distribution $p(\bm{z_c}|Y)$ shown in Equation \ref{Markov}, we use a mixture of Gaussians which has been shown to promote latent identifiability both theoretically and empirically ~\cite{kivva:identifiablity}. We will discuss further details of the conditional prior in Section \ref{sec:implementation}.
Furthermore, as depicted in Figure \ref{fig:vae model}, we use two decoders parameterized by neural networks to jointly model the distribution $p_{\theta}(\bm{x},\tilde{Y}|Y,\bm{z})$ as:
\begin{equation}
	{p_{\theta}(\bm{x},\tilde{Y}|Y,\bm{z})=p_{\theta_1}(\bm{x}|Y,\bm{z}_{c})p_{\theta_2}(\tilde{Y}|\bm{x}, \bm{z})}
\end{equation}
It is worth noting that for the $p_{\theta_2}(\tilde{Y}|\bm{x}, \bm{z})$ is different from the noise transition matrix in the traditional sense, i.e., $p(\tilde{Y}|Y,\bm{x})$. 
Since $Y$ is $d$-separated from $\Tilde{Y}$ conditioned on $\bm{z}$ and $\bm{x}$, $\bm{x}$ and $\bm{z}$ convey sufficient information for deriving $\Tilde{Y}$ \cite{peters:elements}.

For the inference model, we use two variational approximations to jointly model the posterior distribution $q_{\phi}(\bm{z},Y|\bm{x})$, which can be decomposed as:
\begin{equation}
	q_{\phi}(\bm{z},Y|\bm{x})=q_{\phi_1}(Y|\bm{x})q_{\phi_2}(\bm{z}|Y,\bm{x})
\end{equation}
Here, we do not include $\tilde{Y}$ as a conditioning variable in $q_{\phi_1}(Y|\bm{x})$ since the noisy label $\tilde{Y}$ is absent in test data. Therefore, $q_{\phi_1}(Y|\tilde{Y}, \bm{x})$ cannot be directly employed to predict the clean labels in the test set. 
Instead, 
% we assume that given $\bm{x}$, the clean label is conditionally independent of the noisy label. 
we directly use the encoder $q_{\phi_1}(Y|\bm{x})$  to predict the clean labels for given test instance $\bm{x}$.
This design choice does not conflict with our structural causal model and noisy label generation process, because every $\bm{x}$ in datasets should contain enough feature information to predict its true label, and the clean label of each $\bm{x}$ is determined when the instance is generated, regardless of the annotation process or annotators.

An important characteristic that differentiates ReIDVAE from existing semi-supervised and supervised VAE-based methods (e.g., \cite{sohn:learning}) is that labels in ReIDVAE are exploited as auxiliary information instead of directly used as latent variables.
Most of the existing works straightforwardly split the latent space as $\bm{z}=\{\bm{z}_y, \bm{z}_{\backslash y}\}$ and fix $\bm{z}_y={Y}$, and thus $\bm{z}_y$ is forced to directly encoding the label themselves. 
%However, when learning from noisy labels, following the standard setting would cause the algorithm to learn noisy information and cause over-fitting since the clean labels cannot be obtained. 
However, when learning from noisy labels, following the standard setting could lead to learning noisy information and over-fitting, as clean labels are not available.
%To avoid the above problem and eliminate the impact of the directly using of noise labels, 
To avoid the aforementioned problem and eliminate the impact of using labels that may contain noise directly,
we propose to employ the label information as auxiliary information to capture the characteristics. 
To achieve this goal, we also divide $\bm{z}$ into two components such that $\bm{z}=\{\bm{z}_{c}, \bm{z}_{s}\}$. However, $\bm{z}_c$ is designed as the content latent which only captures the characteristics associated with the latent clean labels; $\bm{z}_{s}$ naturally capture style information, which is not causally related to the clean labels, but may affect the generation of noise labels.
The next problem is that the characteristics of different labels will be entangled in $\bm{z}_c$. Therefore, we divide the latent space so that the classification of specific labels $y^i$ can only access the specific latent variable $\bm{z}_c$, which is formulated as $p({\bm{z}_c}|Y) = \sum\nolimits_i {\log p(z_c^i|{y^i})} $.

To efficiently learn the parameters in the encoders and decoders, we utilize amortized variational inference and the re-parameterization trick to optimize the evidence lower-bound ELBO($\bm{x}, \tilde{Y}$) of the marginal likelihood $p_{\theta}(\bm{x},\tilde{Y})$. The ELBO is a lower bound of the likelihood function. 
Briefly, the objective of ReIDVAE can be defined as: 
\allowdisplaybreaks
\begin{align}
    \mathcal{L}_{\text{ELBO}(\bm{x},\tilde{Y})} =&\iint q_{\varphi, \phi}(\boldsymbol{z}, Y \mid \boldsymbol{x}) \log \frac{p(Y) p\left(\boldsymbol{z}_c \mid Y\right) p_{\theta_1}\left(\boldsymbol{x} \mid \boldsymbol{z}_{c}, Y\right) p_{\theta_2}(\tilde{Y} \mid \boldsymbol{x}, \boldsymbol{z})}{q_\phi(\boldsymbol{z}, Y \mid \boldsymbol{x})} d \boldsymbol{z} d Y  \nonumber\\
            =&\iint q_{\phi_1}(Y \mid \boldsymbol{x}) q(\boldsymbol{z} \mid \boldsymbol{x}) \frac{q\left(Y \mid \boldsymbol{x}, \boldsymbol{z}_c\right)}{q(Y \mid \boldsymbol{x})} \nonumber\\
            & \log \frac{p(Y) p\left(\boldsymbol{z}_c \mid Y\right) p_{\theta_1}\left(\boldsymbol{x} \mid \boldsymbol{z}_{c}, Y\right) p_{\theta_2}(\tilde{Y} \mid \boldsymbol{x}, \boldsymbol{z}) q(Y \mid \boldsymbol{x})}{q_{\phi_1}(Y \mid \boldsymbol{x}) q\left(Y \mid \boldsymbol{x}, \boldsymbol{z}_c\right) q(\boldsymbol{z} \mid \boldsymbol{x})} d \boldsymbol{z} d Y \nonumber\\
            =&E_{q_{\phi_1}(Y \mid \bm{x}) q(\bm{z} \mid \bm{x})}\left[\frac{q\left(Y \mid \bm{x}, \bm{z}_c\right)}{q_{\phi_1}(Y \mid \bm{x})} \log \left(\frac{p_{\theta_1}(\bm{x} \mid \bm{z}_{c}, Y) p\left(\bm{z}_c \mid Y\right)}{q\left(Y \mid \bm{x}, \bm{z}_c\right) q(\bm{z} \mid \bm{x})}\right)\right] \nonumber \\
            &+\log p_{\theta_2}(\tilde{Y} \mid \bm{z}, \bm{x})+\log p(Y)
            % \nonumber
    \label{elbo}
\end{align}
%\begin{align}
    %&\mathcal{L}_{\text{ELBO}(\bm{x},\tilde{Y})} \nonumber\\   
            %=&\iint q_{\varphi, \phi}(\boldsymbol{z}, Y \mid \boldsymbol{x}) \nonumber\\ 
            %&\log \frac{p(Y) p\left(\boldsymbol{z}_c \mid Y\right) p_{\theta_1}\left(\boldsymbol{x} \mid \boldsymbol{z}_{c}, Y\right) p_{\theta_2}(\tilde{Y} \mid \boldsymbol{x}, \boldsymbol{z})}{q_\phi(\boldsymbol{z}, Y \mid \boldsymbol{x})} d \boldsymbol{z} d Y  \nonumber\\
            %=&\iint q_{\phi_1}(Y \mid \boldsymbol{x}) q(\boldsymbol{z} \mid \boldsymbol{x}) \frac{q\left(Y \mid \boldsymbol{x}, \boldsymbol{z}_c\right)}{q(Y \mid \boldsymbol{x})} \nonumber\\
            %& \log \frac{p(Y) p\left(\boldsymbol{z}_c \mid Y\right) p_{\theta_1}\left(\boldsymbol{x} \mid \boldsymbol{z}_{c}, Y\right) p_{\theta_2}(\tilde{Y} \mid \boldsymbol{x}, \boldsymbol{z}) q(Y \mid \boldsymbol{x})}{q_{\phi_1}(Y \mid \boldsymbol{x}) q\left(Y \mid \boldsymbol{x}, \boldsymbol{z}_c\right) q(\boldsymbol{z} \mid \boldsymbol{x})} d \boldsymbol{z} d Y \nonumber\\
            %=&E_{q_{\phi_1}(Y \mid \bm{x}) q(\bm{z} \mid \bm{x})}\left[\frac{q\left(Y \mid \bm{x}, \bm{z}_c\right)}{q_{\phi_1}(Y \mid \bm{x})} \log \left(\frac{p_{\theta_1}(\bm{x} \mid \bm{z}_{c}, Y) p\left(\bm{z}_c \mid Y\right)}{q\left(Y \mid \bm{x}, \bm{z}_c\right) q(\bm{z} \mid \bm{x})}\right)\right] \nonumber\\
            %&+\log p_{\theta_2}(\tilde{Y} \mid \bm{z}, \bm{x})+\log p(Y)
            % \nonumber
%\end{align}
It is worth noting that the classifier term in both the training and test procedure, i.e. $p_{\theta_1}(\tilde{Y} \mid \bm{z}, \bm{x})$, $q_{\phi_1}(Y \mid \bm{x})$, appear naturally from the structural causal model and ELBO derivation. For complete ELBO derivation and other items in this objective function, please  refer to the Appendices.
%For complete ELBO derivation, please  refer to the Appendices. It is worth noting that the classifier term in both the training and test procedure, i.e. $p_{\theta_1}(\tilde{Y} \mid \bm{z}, \bm{x})$, $q_{\phi_1}(Y \mid \bm{x})$, appear naturally from the causal model and ELBO derivation. 
%In additional to the encoders and decoders parameterized by DNN, other items in this objective function are described as follows: the approximate posterior is designed as $q(\bm{z}|\bm{x})=\mathcal{N}(\bm{z}_{c}, \bm{z}_{s}|\mu(\bm{x}), diag(\sigma^{2}(\bm{x})))$, where $\mu(\bm{x})$ and $diag(\sigma^{2}(\bm{x}))$ is the architecture from  ~\cite{burgess:understanding}; the label predictive distribution $q(Y|\bm{x}, \bm{z}_{c})$ is represented as $Bernoulli(Y|\pi(\bm{z}_{c}),\bm{x})$, where $\pi(\bm{z}_{c})$ is a diagonal transformation forcing the factorization $q(Y|\bm{x},\bm{z}_{c})=\prod_{i}{q(Y_i|\bm{z}_c^i,\bm{x}_i)}$; The conditional prior is designed as $p(\bm{z}_{c}|Y)=\mathcal{N}(\bm{z}_{c}|\mu(Y), diag(\sigma^{2}(Y)))$, where the parameters are parameterized by neural network.
\subsection{Conditional Mixture Prior}\label{sec:implementation}
Inspired by the identifiability \cite{kivva:identifiablity} of VAE with mixtures priors distributions, we design our model with a novel way to integrate co-teaching to infer two conditional prior distributions.
%Traditionally, co-teaching has been used to increase the robustness of the deep models for learning from noisy labels by the small loss heuristics, however, in ReIDVAE the two branches are designed to learn two components for the mixture of latent prior distribution. 
% The two branches work in the same way, so only the workflow of the first branch is introduced.

Figure \ref{fig:vae model} illustrates a branch of our model for learning one component of the mixture. Compared with previous methods, ReIDVAE not only excludes $\bm{z}_s$ and reconstructs the style-free $\hat{\bm{x}}$, but also utilizes the label characteristics encoded in $\bm{z}$ instead of $Y$ when decoding the noisy labels. 
When feeding an instance $\bm{x}$ to the model, encoder $q_{\phi_{1}^1}(Y|\bm{x})$ outputs a predicted clean label $Y$. 
Utilizing both the instance $\bm{x}$ and the predicted clean label, encoder $q_{\phi_{2}^{1}}(\bm{z}|Y,\bm{x})$ outputs the latent representation $\bm{z}=\{\bm{z}_{s}, \bm{z}_{c} \}$ that contains both style and content information. 
In order to reduce the influence of latent features irrelevant the class label when predicting the clean label, the style latent $\bm{z}_{s}$ is then discarded when calculating the loss with respect to the predicted clean label $Y$. 
The content latent $\bm{z}_{c}$ and the predicted clean class label $Y$ are then used by $p_{\theta_{1}^1}(\bm{x}|Y, \bm{z}_{c})$ to reconstruct the style-free image $\hat{\bm{x}}$. 
Finally, the reconstructed $\hat{\bm{x}}$ and the full latent $\bm{z}=\{\bm{z}_{s}, \bm{z}_{c} \}$ are utilized by $p_{\theta_{2}^1}(\tilde{Y}|\hat{\bm{x}}, \bm{z})$ to decode the noisy label $\tilde{Y}$. 

The above process has two major benefits. Firstly, 
% In this process, on the one hand, the reconstructed $\hat{\bm{x}}$ excludes the style information of $\bm{x}$ and only contains the content. 
%since the reconstructed $\hat{\bm{x}}$ excludes the style latent $\bm{z}_s$, it is easier to for our model to predict the clean label $Y$ as $\bm{z}_s$ is a cause to instance-dependent label noise. 
since the reconstructed $\hat{\bm{x}}$ excludes the style latent $\bm{z}_s$, our designed reconstruction process, i.e., $p_{\theta_{1}^1}(\bm{x}|Y, \bm{z}_{c})$,  further promotes learning of $\bm{z}_c$ and estimation of $Y$ by compelling $\hat{\bm{x}}$ which contains only category information to be as similar as possible to the original $\bm{x}$.
Secondly, since both the style and content features affect the generation of noisy labels, the full latent representation $\bm{z}=\{\bm{z}_c,\bm{z}_s\}$ is utilized in conjunction with $\hat{\bm{x}}$ to predict the noisy label $\tilde{Y}$. 
This allows our work to sidestep the challenge of dealing with the high dimensional $\bm{x}$ and the unavailability of clean labels $Y$. Instead, we focus on the inferred representation $\bm{z}$, which has a lower dimensionality than $\bm{x}$ and more substantial representational power than $Y$. 
By utilizing $\bm{z}$ as the determining factor, we can mitigate the risks associated with directly using the estimated $Y$, especially when the classifier's performance is poor during the early stages of training.

The workflow of the encoders $q_{\phi_{1}^2}(Y|\bm{x})$, $q_{\phi_{2}^{2}}(\bm{z}|Y,\bm{x})$ and decoders $p_{\theta_{1}^2}(\bm{x}|Y,\bm{z}_{c})$, $p_{\theta_{2}^2}(\tilde{Y}|\hat{\bm{x}},\bm{z})$ of the second branch is the same as the above process. 
When compared with traditional label noise learning approaches where co-teaching is often used with small-loss heuristic for selecting samples with clean labels or increasing the robustness of the model, in our approach the two co-teaching branches have fundamentally different utilities, each of the two branches of ReIDVAE infers a conditional latent prior distribution, and the two prior distributions are tied together with the co-teaching loss function. Therefore, the two branches of the latent conditional priors effectively act as two components of the mixture prior conditional distribution, and thus benefit the learning procedure by improving training stability and latent identifiability \cite{kivva:identifiablity}.

\begin{table*}[h]
    \small
	\begin{center}
		\caption{Means and standard deviations (percentage) of classification accuracy on FashionMNIST with different label noise levels.\\}
		%\begin{tabular}{llllll}	
		\begin{tabular}{lccccc}	
			\hline
							& IDN-20\%                    & IDN-30\%                   &   IDN-40\%                &  IDN-45\%            & IDN-50\%            \\ \hline
			CE               & 88.54$\pm$0.32             & 88.38$\pm$0.42             &   84.22$\pm$0.35          &  69.72$\pm$0.72      & 52.32$\pm$0.68          \\
			Co-teaching      & 91.21$\pm$0.31             & 90.30$\pm$0.42             &   89.10$\pm$0.29          &  86.78$\pm$0.90      & 63.22$\pm$1.56          \\
			Decoupling       & 90.70$\pm$0.28             & 90.34$\pm$0.36             &   88.78$\pm$0.44          &  87.54$\pm$0.53      & 68.32$\pm$1.77          \\
			Mixup            & 88.68$\pm$0.37             & 88.02$\pm$0.37             &   85.47$\pm$0.55          &  79.57$\pm$0.75      & 66.02$\pm$2.58          \\
			Forward          & 90.05$\pm$0.43             & 88.65$\pm$0.43             &   86.27$\pm$0.48          &  73.35$\pm$1.03      & 58.23$\pm$3.14          \\
			Reweight         & 90.27$\pm$0.27             & 89.58$\pm$0.37             &   87.04$\pm$0.32          &  80.69$\pm$0.89      & 64.13$\pm$1.23          \\
			T-Revision       & 91.58$\pm$0.31             & 90.11$\pm$0.61             &   89.46$\pm$0.42          &  84.01$\pm$1.14      & 68.99$\pm$1.04          \\ \hline
            PTD              & 89.78$\pm$0.43             & 88.30$\pm$0.51             &   80.75$\pm$2.86          &  80.17$\pm$0.15      & 72.22$\pm$4.22          \\
            BLTM             & 89.70$\pm$0.14             & 87.63$\pm$1.87             &   78.40$\pm$3.69          &  79.41$\pm$0.16      & 66.55$\pm$7.52          \\
			CausalNL			& 90.84$\pm$0.31             & 90.68$\pm$0.37             &   90.01$\pm$0.45          &  88.75$\pm$0.81      & 78.19$\pm$1.01          \\ \hline
			ReIDVAE          & \textbf{92.50$\pm$0.09}    & \textbf{91.68$\pm$0.11}    &   \textbf{90.94$\pm$0.12} &  \textbf{88.87$\pm$0.75}      & \textbf{84.92$\pm$0.19}   \\ \hline
			
		\end{tabular}
	\end{center}	
\end{table*}

\begin{table*}[!h]
    \small
	\begin{center}
		\caption{Means and standard deviations (percentage) of classification accuracy on SVHN with different label noise levels.\\}
		%\begin{tabular}{llllll}	
		\begin{tabular}{lccccc}	
			\hline  
							& IDN-20\%                  & IDN-30\%                  &   IDN-40\%               &  IDN-45\%            & IDN-50\%            \\ \hline
			CE               & 91.51$\pm$0.45           & 91.21$\pm$0.43            &   87.87$\pm$1.12         &  67.15$\pm$1.65      & 51.01$\pm$3.62          \\
			Co-teaching      & 93.93$\pm$0.31           & 92.06$\pm$0.31            &   91.93$\pm$0.81         &  89.33$\pm$0.71      & 67.62$\pm$1.99          \\
			Decoupling       & 90.02$\pm$0.25           & 91.59$\pm$0.25            &   88.27$\pm$0.42         &  84.57$\pm$0.89      & 65.14$\pm$2.79          \\
			Mixup            & 89.73$\pm$0.37           & 90.02$\pm$0.35            &   85.47$\pm$0.63         &  82.41$\pm$0.62      & 68.95$\pm$2.58          \\
			Forward          & 91.89$\pm$0.31           & 91.59$\pm$0.23            &   89.33$\pm$0.53         &  80.15$\pm$1.91      & 62.53$\pm$3.35          \\
			Reweight         & 92.44$\pm$0.34           & 92.32$\pm$0.51            &   91.31$\pm$0.67         &  85.93$\pm$0.84      & 64.13$\pm$3.75          \\
			T-Revision       & 93.14$\pm$0.53           & 93.51$\pm$0.74            &   92.65$\pm$0.76         &  88.54$\pm$1.58      & 64.51$\pm$3.42          \\ \hline
            PTD              & 94.19$\pm$0.20           & 92.56$\pm$0.83            &   88.13$\pm$1.56         &  78.38$\pm$0.03      & 77.04$\pm$2.56          \\
            BLTM             & \textbf{94.83$\pm$0.64}  & 92.43$\pm$0.91            &   86.91$\pm$1.17         &  64.99$\pm$0.75      & 76.53$\pm$2.15          \\
			CausalNL		 & 94.06$\pm$0.23           & 93.86$\pm$0.65            &  \textbf{93.82$\pm$0.64} &  93.19$\pm$0.93      & 85.41$\pm$2.95          \\ \hline
			ReIDVAE          & 94.38$\pm$0.08           & \textbf{94.08$\pm$0.09}   & 93.43$\pm$0.11           &  \textbf{93.29$\pm$0.64} & \textbf{92.96$\pm$0.11}   \\ \hline
			
		\end{tabular}
	\end{center}	
\end{table*}

Finally, the loss function of ReIDVAE can be expressed as:
\begin{equation}
	\mathcal{L}_{total}=\ -\text{ELBO}(\bm{x},\ \widetilde{Y})+\alpha \mathcal{L}_{\text{co-teaching}}
\end{equation}
where the $\mathcal{L}_{\text{co-teaching}}$ is the co-teaching loss and $\alpha$ is a weighting parameter. The ELBO presented in Equation \ref{elbo} comprises the reconstruction error between the $\hat{\bm{x}}$ output by the $p_{\theta_{1}}(\bm{x}|Y,\bm{z}_{c})$ and the original $\bm{x}$, as well as the cross entropy loss between the noise label $\tilde{y}$ produced by the second decoder, i.e., $p_{\theta_{2}}(\tilde{Y}|\hat{\bm{x}},\bm{z})$ and the $\tilde{Y}$ in training set.

%where the $\mathcal{L}_{\text{co-teaching}}$ is the co-teaching loss and $\alpha$ is a weighting parameter. The ELBO shown in Equation \ref{elbo} contains the reconstruction error between the $\hat{\bm{x}}$ output by first decoder, i.e., $p_{\theta_{1}^1}(\bm{x}|Y,\bm{z}_{c})$ and the original $\bm{x}$, and the cross entropy loss between the noise label $\tilde{y}$ output by the second decoder, i.e., $p_{\theta_{2}^1}(\tilde{Y}|\bm{x},\bm{z})$ and the noise $\tilde{Y}$ in training set. \comment{The grammer of this sentence is not right. Also it's bad to have one sentence with two "i.e.". }

\section{EXPERIMENTS}
\subsection{Experiment Setup}

\textbf{Datasets} 
We evaluate our proposed ReIDVAE method on a wide range of benchmark datasets with both controlled and real-world instance-dependent noisy labels. 
Firstly, we follow the IDN generation process in ~\cite{xia:part} and corrupt the FashionMNIST~\cite{xiao:fashion}, SVHN~\cite{netzer:reading}, CIFAR-10 and CIFAR-100~\cite{krizhevsky:learning} datasets with various amounts of noisy labels. 
Specifically, FashionMNIST contains 60,000 training images and 10,000 test images with 10 classes; SVHN contains 73,257 training images and 26,032 test images with 10 classes; CIFAR-10 and CIFAR-100 contain 50,000 training images and 10,000 test images of size 32*32 with 10 and 100 classes, respectively.
Secondly, we evaluate ReIDVAE against baselines on a webly-supervised and a human-annotated real-world instance-dependent noise datasets Clothing1M~\cite{xiao:learning} and CIFAR-10N~\cite{wei:learning}
%\note{Please fix the missing reference}.
The training set of Clothing1M contains 1M clothing images with 14 classes of webly-supervised real-world noisy labels, and the test set consists of 10k samples with hand-selected clean labels. CIFAR-10N contains 5 different levels of instance-dependent noisy labels provided by human annotators from Amazon Mechanical Turk.

\begin{table*}[!h]
    \small
	\begin{center}
		\caption{Means and standard deviations (percentage) of classification accuracy on CIFAR-10 with different label noise levels.\\}
		%\begin{tabular}{lllllll}	
		\begin{tabular}{lcccccc}
			\hline
			             & IDN-20\%                 & IDN-30\%                 &   IDN-40\%                &  IDN-45\%             &   IDN-50\%                &   IDN-60\%     \\ \hline
			CE           & 79.40$\pm$0.37           & 70.63$\pm$5.25           &   60.93$\pm$2.71         &  54.80$\pm$4.79         &   49.06$\pm$3.80          &   37.27$\pm$1.48        \\
			Co-teaching  & 86.77$\pm$0.09           & 83.83$\pm$0.49           &   80.28$\pm$0.88         &  75.29$\pm$0.56         &   55.08$\pm$1.61          &   26.84$\pm$0.41        \\
			Decoupling   & 81.97$\pm$0.58           & 73.55$\pm$1.40           &   61.74$\pm$2.19         &  54.57$\pm$2.56         &   49.81$\pm$0.94          &   36.50$\pm$1.95   \\
			Mixup        & 83.19$\pm$2.49           & 75.51$\pm$2.84           &   64.83$\pm$2.08         &  58.73$\pm$1.88         &   52.49$\pm$4.25          &   39.45$\pm$2.77     \\
			Forward      & 78.13$\pm$0.04           & 75.78$\pm$0.01           &   68.75$\pm$0.08         &  54.69$\pm$0.46         &   45.85$\pm$0.36          &   27.19$\pm$0.09        \\
			Reweight     & 79.15$\pm$0.07           & 71.01$\pm$0.05           &   63.01$\pm$0.06         &  57.52$\pm$0.19         &   55.44$\pm$0.31          &   42.95$\pm$0.21        \\
			T-Revision   & 78.31$\pm$0.02           & 76.92$\pm$0.04           &   69.19$\pm$0.06         &  63.36$\pm$0.05         &   57.42$\pm$0.05          &   44.14$\pm$0.01        \\
            \hline
            PTD          & \textbf{88.46$\pm$0.13}  & 85.49$\pm$0.27           &   83.92$\pm$0.12         &  83.06$\pm$0.19         &   73.97$\pm$0.19          &   63.41$\pm$0.17        \\
            BLTM         & 87.91$\pm$0.09           & \textbf{87.75$\pm$0.10}  &   77.36$\pm$0.16         &  76.79$\pm$0.13         &   74.66$\pm$0.18          &   46.96$\pm$0.25        \\
			CausalNL     & 86.50$\pm$0.16           & 83.48$\pm$0.90           &   79.63$\pm$2.01         &  74.43$\pm$2.06         &   64.40$\pm$3.03          &   40.93$\pm$4.00        \\ \hline
			ReIDVAE      & 87.15$\pm$0.15           & 85.48$\pm$0.38           &  \textbf{85.07$\pm$0.32} & \textbf{84.15$\pm$1.02} &  \textbf{79.08$\pm$0.68}  &   \textbf{75.27$\pm$0.83} \\ \hline
			
		\end{tabular}
	\end{center}	
\end{table*}

\begin{table*}[!h]
    \small
	\begin{center}
		\caption{Means and standard deviations (percentage) of classification accuracy on CIFAR-100 with different label noise levels. \\}
		%\begin{tabular}{lllllll}	
		\begin{tabular}{lcccccc}
			\hline
			             & IDN-20\%                 & IDN-30\%                 &   IDN-40\%               &  IDN-45\%              &   IDN-50\%              &   IDN-60\%     \\ \hline
		    CE           & 56.86$\pm$0.35           & 49.85$\pm$0.72           &   41.95$\pm$0.54         &  37.90$\pm$0.51        &   34.45$\pm$0.33        &   25.91$\pm$0.27        \\
			Co-teaching  & 64.15$\pm$0.41           & 59.50$\pm$0.30           &   54.50$\pm$0.33         &  49.67$\pm$0.60        &   44.62$\pm$0.51        &   30.53$\pm$0.31        \\
			Decoupling   & 55.85$\pm$0.42           & 49.99$\pm$0.62           &   41.98$\pm$0.64         &  38.41$\pm$0.49        &   34.18$\pm$0.98        &   26.38$\pm$0.56   \\
			Mixup        & 58.81$\pm$2.00           & 53.51$\pm$1.79           &   45.28$\pm$1.67         &  40.90$\pm$2.24        &   37.80$\pm$1.19        &   32.73$\pm$2.57     \\
			Forward      & 59.27$\pm$0.02           & 52.56$\pm$0.01           &   44.14$\pm$0.02         &  41.71$\pm$0.01        &   37.23$\pm$0.01        &   28.14$\pm$0.02        \\
			Reweight     & 58.59$\pm$0.05           & 51.85$\pm$0.02           &   44.39$\pm$0.02         &  41.10$\pm$0.02        &   36.22$\pm$0.05        &   28.20$\pm$0.03        \\
			T-Revision   & 59.00$\pm$0.01           & 52.95$\pm$0.03           &   47.27$\pm$0.06         &  43.78$\pm$0.03        &   39.85$\pm$0.06        &   28.66$\pm$0.07        \\
            \hline
            PTD          & 33.19$\pm$0.01           & 19.48$\pm$0.04           &   20.87$\pm$0.01         &  11.57$\pm$0.05        &   11.92$\pm$0.01        &   2.49$\pm$0.02        \\  
            BLTM         & 59.85$\pm$0.15           & 56.84$\pm$0.17           &   47.30$\pm$0.26         &  42.52$\pm$0.48        &   37.67$\pm$0.32        &   24.99$\pm$0.34        \\
			CausalNL     & 67.35$\pm$0.40           & 65.93$\pm$0.23           &   63.34$\pm$0.07         &  61.14$\pm$0.07        &  57.15$\pm$0.27        &   39.69$\pm$1.24        \\ \hline
			ReIDVAE      & \textbf{71.57$\pm$0.04}      & \textbf{70.72$\pm$0.09}      &   \textbf{69.40$\pm$0.06}    & \textbf{66.93$\pm$0.08}    &  \textbf{64.00$\pm$0.10}    &  \textbf{51.90$\pm$0.08} \\ \hline
			
		\end{tabular}
	\end{center}	
\end{table*}

\noindent\textbf{Baselines} We compare ReIDVAE with a wide range of baselines including both non-IDN and IDN methods. 
For non-IDN methods, we compare with: 
(i) Cross Entropy(CE), which trains a standard deep network with the cross-entropy loss on the noisy datasets;
(ii) Co-teaching~\cite{han:co}, which handles noisy labels by training two networks using the small heuristics;
(iii) Decoupling~\cite{malach:decoupling}, which trains two networks on samples whose predictions from the two networks are different;
(iv) Mixup~\cite{zhang:mixup}, which straightforwardly interpolates the training samples and labels; 
(v) Forward~\cite{patrini:forward}, Re-weight~\cite{liu:classification}, and T-Revision~\cite{xia:anchor}, which adopt different approaches to utilize class-dependent transition matrix for correcting the loss function.
For IDN methods, we compare with:
(vi) PTD~\cite{xia:part} which assumes the noisy labels are part-dependent; 
(vii) BLTM~\cite{yang:estimating}, which aims to estimate the instance-dependent noise transition matrix; 
(vii) CausalNL~\cite{yao:instance}, which utilizes a structural causal model to address IDN.

\noindent\textbf{Implementation}
For a fair comparison, we conduct all experiments on NVIDIA GeForce RTX 3090 GPUs, and all methods are implemented with PyTorch 1.10.0. 
%\note{This cannot be right, PyTorch has no version 3.9}. 
We use a ResNet-18 network for FashionMNIST and a ResNet-34 network for SVHN.
% , the dimension of the latent representation $Z$ is set to 25.
%, and the optimization strategy we used is Adam with a learning rate $10^{-3}$ on both datasets. 
For CIFAR-10 and CIFAR-100, two wide-ResNet~\cite{zagoruyko:wide} are utilized as backbone.
%Adam is adopted as the optimizer with learning rate $10^{-3}$ and $10^{-4}$, respectively. 
For Clothing1M, we use ResNet-50 with pre-trained weights. 
For further implementation and parameter details, please refer to the supplementary materials.

\subsection{Results}
\noindent\textbf{Controlled IDN}
%\note{PTD results is missing from Table 4!!.}
Tables 1,2,3 and 4 report the classification accuracy of the datasets with controlled instance-dependent label noise including FashionMNIST, SVHN, CIFAR-10 and CIFAR-100, respectively. The controlled IDN experiments reveal that our ReIDVAE method achieves better performances than the compared baselines in almost all cases.
Furthermore, as the noise rate increases, the performances of ReIDVAE decrease much slower than the baselines. 
%Most importantly, when the noise rate is high, e.g., greater than 40\%,  ReIDVAE can still perform quite well, while all baseline performances degenerate significantly.

\begin{table*}[!h]
    \small
	\begin{center}
		\caption{Means and standard deviations (percentage) of classification accuracy on CIFAR-10N with different label noise levels.\\}
		%\begin{tabular}{llllll}
        \begin{tabular}{lccccc}	
			\hline
			                 %& Random1 (17.23$\%$)      & Random2 (18.12$\%$)      &   Random3 (17.64$\%$)     &  Aggregate (9.03$\%$)   &  Worst (40.21$\%$)         \\ \hline
                              & Random1                  & Random2                  &   Random3                 &  Aggregate             &  Worst \\
                              &(17.23$\%$)               & (18.12$\%$)              &   (17.64$\%$)             &  (9.03$\%$)            &  (40.21$\%$)
                              \\ \hline
                CE            & 81.58$\pm$0.89           & 81.54$\pm$0.40           &   81.84$\pm$0.40          &  86.84$\pm$0.15        &  63.10$\pm$0.85               \\
                Co-teaching   & 88.47$\pm$0.15           & 88.55$\pm$0.08           &   88.26$\pm$0.15          &  89.91$\pm$0.04        &  81.84$\pm$0.17               \\
                Decoupling    & 82.71$\pm$0.34           & 81.48$\pm$0.56           &   82.21$\pm$0.44          &  87.01$\pm$0.14        &  64.90$\pm$1.17               \\
                Mixup         & 83.55$\pm$0.07           & 82.73$\pm$0.07           &   81.22$\pm$0.03          &  85.85$\pm$0.03        &  62.27$\pm$2.52               \\
                Forward       & 85.83$\pm$0.09           & 84.57$\pm$0.09           &   84.79$\pm$0.02          &  86.62$\pm$0.01        &  76.96$\pm$0.07               \\
                Reweight      & 82.12$\pm$0.08           & 82.34$\pm$0.02           &   82.16$\pm$0.02          &  87.52$\pm$0.07        &  66.82$\pm$0.06               \\
                T-Revision    & 82.40$\pm$0.06           & 83.03$\pm$0.06           &   80.31$\pm$0.05          &  85.53$\pm$0.01        &  70.44$\pm$0.07               \\
            \hline
                PTD           & 89.41$\pm$0.11           & 88.71$\pm$0.05           &   89.10$\pm$0.13          &  90.11$\pm$0.08        &  81.89$\pm$0.21               \\
                BLTM          & 87.32$\pm$0.20           & 86.13$\pm$0.25           &   87.29$\pm$0.19          &  88.85$\pm$0.11        &  80.69$\pm$0.17               \\
			  CausalNL      & 84.18$\pm$0.16           & 83.86$\pm$0.90           &   84.22$\pm$2.01          &  87.73$\pm$2.06        &  64.50$\pm$3.03                \\ \hline
			  ReIDVAE       & \textbf{92.54$\pm$0.01}      & \textbf{92.53$\pm$0.01}      &   \textbf{92.55$\pm$0.01}     & \textbf{93.74$\pm$0.02}    &  \textbf{86.35$\pm$0.10}           \\ \hline
			
		\end{tabular}
	\end{center}	
\end{table*}

\begin{table*}[!h]
    \small
    \setlength{\tabcolsep}{1mm}
	\begin{center}
		\caption{Classification accuracy on Clothing1M. In the experiments, only noisy samples are exploited during the training process.\\}
		%\begin{tabular}{lllllllll}
        \begin{tabular}{cccccccccccc}
			\hline
			%CE    & Decoupling  & MentorNet & Co-teaching & Forward & Reweight & T-Revision & causalNL & VAE\_wide \\ \hline
			%68.88 & 54.53       & 56.79     & 60.15       & 69.91   & 70.40    & 70.97      & 72.24    & $\bm{76.5533}$   \\ \hline
            %CE    & Co-teaching & Decoupling & Forward & Reweight & T-Revision  & PTD    & BLTM   & causalNL & ReIDVAE \\ \hline
			%68.88 & 60.15       & 54.53      & 69.91   & 70.40    & 70.97       & 71.67  & 73.39   & 72.24    & \textbf{76.55}  \\ \hline
            CE    & Co-teaching & Decoupling & Forward & Reweight  &T-Revision  & PTD    & BLTM   & causalNL & ReIDVAE\\ \hline
            68.88 & 60.15       & 54.53      & 69.91   & 70.40     &70.97       & 71.67  & 73.39   & 72.24    & \textbf{76.55}\\ \hline
            %T-Revision  & PTD    & BLTM   & causalNL & ReIDVAE \\ \hline
            %70.97       & 71.67  & 73.39   & 72.24    & \textbf{76.55}  \\ \hline
		\end{tabular}
	\end{center}
\end{table*}

An important advantage of ReIDVAE is that our method achieves great performances when the noise rate is high (e.g., equal or greater than 40\%) and can perform well even at 60\% of noises, which has never been achieved by previous methods. %\note{It may be better to discuss Wide-ResNet backbone in the supplementary.}
For example, on FashionMNIST and SVHN, when the noise rate increases from 20\% to 50\%, the accuracies of our method only decrease by 7.58\% and 1.42\%. However, the performances of the compared baselines degenerate significantly and their classification accuracies are at least 6\% lower than ReIDVAE when the noise rates are at 50\%.
From Table 3 and Table 4, we can see that the high noise rate performances on CIFAR-10 and CIFAR-100 are better than the baselines. 
In the hardest cases when the instance-dependent noise rate exceeds 50\% and reaches 60\% on CIFAR10 and CIFAR100, our method can still reach 75.27\% and 51.9\% accuracy, which are more than 10\% ahead of the closest contending baselines. 
This evidence supports the fact that ReIDVAE successfully considers the instance-dependent noise generation process and exploits the weak supervision information in the datasets, and thus the accuracies of our method decrease slowly with the increase of the noise rates. 

In addition, although ReIDVAE still performs significantly better than the compared baselines on CIFAR-100, we observe that the performances of ReIDVAE decrease faster on CIFAR-100 than on other datasets as the noise rate gradually increases to 60\%.
This is because the number of categories is relatively large when compared to the number of samples. As we need to separate the content latent variable related to each category to capture the unique characteristics of each category, the dimension of content latents will increase with the number of categories; therefore, it would be more difficult to differentiate among each component of $\bm{z}_c$. 
%However, even on datasets that are unfavorable to our algorithm, our method still achieves state-of-the-art results. 
However, other methods that are sensitive to the number of categories, such as PTD, degrade more severely on this unfavorable dataset.

\noindent\textbf{Real-world IDN}
Tables 5 and 6 report the classification performances on datasets with real-world IDN including CIFAR-10N and Clothing1M. 
We can clearly see that our ReIDVAE achieves significantly better performances in all noise scenarios on CIFAR-10N and Clothing1M.
On CIFAR-10N, it is also worth noting that the performances of many baseline methods degenerate significantly on the \emph{worst} case with less than 50\% of noisy labels, possibly due to their strong assumptions on the noisy label generation process do not align with real-world scenarios.
However, because our ReIDVAE successfully considers the annotation process of real-world annotators, our method achieves state-of-art performance on these real-world noisy datasets. Specifically, without the aid of complex data augmentation and semi-supervised learning, ReIDVAE achieves the classification accuracy of 86\% and 76.55\% on the worst case of CIFAR10-N and Clothing1M, respectively.

\section{Conclusion}
In this paper, we have investigated how to utilize the weak supervision signal conveyed in the instance-dependent noisy labels to infer disentangled latent causal representations for better modeling the instance-dependent label noise generation process.
By inferring low-dimensional latent representations from raw features supervised with noisy labels and disentangling the latents into the content and style components,
we have successfully approximated the causal structure model for generating the instance-dependent noisy datasets.
This has allowed us to encapsulate and disentangle the characteristics associated with labels rather than just the label themselves, which have in turn facilitated the noisy label learning process.
Extensive experimental results on both controlled and real-world noisy datasets demonstrate the effectiveness of our method against various baselines. 

\bibliographystyle{plain}
\bibliography{ref}

\begin{thebibliography}{10}

\bibitem{arpit:closer}
Devansh Arpit, Stanis{\l}aw Jastrz{{e}}bski, Nicolas Ballas, David Krueger,
  Emmanuel Bengio, Maxinder~S. Kanwal, Tegan Maharaj, Asja Fischer, Aaron
  Courville, Yoshua Bengio, and Simon Lacoste-Julien.
\newblock A closer look at memorization in deep networks.
\newblock In {\em Proceedings of the 34th International Conference on Machine
  Learning}, pages 233--242. PMLR, 2017.

\bibitem{berthon:confidence}
Antonin Berthon, Bo~Han, Gang Niu, Tongliang Liu, and Masashi Sugiyama.
\newblock Confidence scores make instance-dependent label-noise learning
  possible.
\newblock In {\em Proceedings of the 38th International Conference on Machine
  Learning}, pages 825--836. PMLR, 2021.

\bibitem{chen:beyond}
Pengfei Chen, Junjie Ye, Guangyong Chen, Jingwei Zhao, and Pheng-Ann Heng.
\newblock Beyond class-conditional assumption: A primary attempt to combat
  instance-dependent label noise.
\newblock In {\em Proceedings of the 35th AAAI Conference on Artificial
  Intelligence}, pages 11442--11450, 2021.

\bibitem{cheng:instance}
De~Cheng, Tongliang Liu, Yixiong Ning, Nannan Wang, Bo~Han, Gang Niu, Xinbo
  Gao, and Masashi Sugiyama.
\newblock Instance-dependent label-noise learning with manifold-regularized
  transition matrix estimation.
\newblock In {\em Proceedings of the IEEE/CVF Conference on Computer Vision and
  Pattern Recognition}, pages 16630--16639, 2022.

\bibitem{cheng:learning}
Hao Cheng, Zhaowei Zhu, Xingyu Li, Yifei Gong, Xing Sun, and Yang Liu.
\newblock Learning with instance-dependent label noise: A sample sieve
  approach.
\newblock {\em arXiv preprint arXiv:2010.02347}, 2020.

\bibitem{cothey:web}
Viv Cothey.
\newblock Web-crawling reliability.
\newblock {\em Journal of the American Society for Information Science and
  Technology}, 55(14):1228--1238, 2004.

\bibitem{han:masking}
Bo~Han, Jiangchao Yao, Gang Niu, Mingyuan Zhou, Ivor Tsang, Ya~Zhang, and
  Masashi Sugiyama.
\newblock Masking: A new perspective of noisy supervision.
\newblock In {\em Advances in Neural Information Processing Systems 31}, page
  5841–5851, 2018.

\bibitem{han:co}
Bo~Han, Quanming Yao, Xingrui Yu, Gang Niu, Miao Xu, Weihua Hu, Ivor Tsang, and
  Masashi Sugiyama.
\newblock Co-teaching: Robust training of deep neural networks with extremely
  noisy labels.
\newblock In {\em Advances in Neural Information Processing Systems 31}, pages
  8536--8546, 2018.

\bibitem{irvin:chexpert}
Jeremy Irvin, Pranav Rajpurkar, Michael Ko, Yifan Yu, Silviana Ciurea-Ilcus,
  Chris Chute, Henrik Marklund, Behzad Haghgoo, Robyn Ball, Katie Shpanskaya,
  et~al.
\newblock Chexpert: A large chest radiograph dataset with uncertainty labels
  and expert comparison.
\newblock In {\em Proceedings of the 33rd AAAI Conference on Artificial
  Intelligence}, pages 590--597, 2019.

\bibitem{jiang:beyond}
Lu~Jiang, Di~Huang, Mason Liu, and Weilong Yang.
\newblock Beyond synthetic noise: Deep learning on controlled noisy labels.
\newblock In {\em Proceedings of the 37th International Conference on Machine
  Learning}, pages 4804--4815. PMLR, 2020.

\bibitem{joy:capturing}
Tom Joy, Sebastian~M Schmon, Philip~HS Torr, N~Siddharth, and Tom Rainforth.
\newblock Capturing label characteristics in vaes.
\newblock {\em In arXiv preprint arXiv:2006.10102}, 2020.

\bibitem{karimi:deep}
Davood Karimi, Haoran Dou, Simon~K Warfield, and Ali Gholipour.
\newblock Deep learning with noisy labels: Exploring techniques and remedies in
  medical image analysis.
\newblock {\em Medical Image Analysis}, 65:101759, 2020.

\bibitem{kingma:vae}
Diederik~P Kingma and Max Welling.
\newblock Auto-encoding variational bayes.
\newblock {\em arXiv preprint arXiv:1312.6114}, 2014.

\bibitem{kivva:identifiablity}
Bohdan Kivva, Goutham Rajendran, Pradeep Ravikumar, and Bryon Aragam.
\newblock Identifiability of deep generative models under mixture priors
  without auxiliary information.
\newblock In {\em Advances in Neural Information Processing Systems 35}, 2022.

\bibitem{krizhevsky:learning}
Alex Krizhevsky, Geoffrey Hinton, et~al.
\newblock Learning multiple layers of features from tiny images.
\newblock {\em Master's thesis, University of Tronto}, 2009.

\bibitem{li:dividemix}
Junnan Li, Richard Socher, and Steven~CH Hoi.
\newblock Dividemix: Learning with noisy labels as semi-supervised learning.
\newblock In {\em Proceedings of the 8th International Conference on Learning
  Representations}, 2020.

\bibitem{liu:early}
Sheng Liu, Jonathan Niles-Weed, Narges Razavian, and Carlos Fernandez-Granda.
\newblock Early-learning regularization prevents memorization of noisy labels.
\newblock In {\em Advances in Neural Information Processing Systems 33}, pages
  20331--20342, 2020.

\bibitem{liu:classification}
Tongliang Liu and Dacheng Tao.
\newblock Classification with noisy labels by importance reweighting.
\newblock {\em IEEE Transactions on Pattern Analysis and Machine Intelligence},
  38(3):447--461, 2015.

\bibitem{locatello:challenging}
Francesco Locatello, Stefan Bauer, Mario Lucic, Gunnar Raetsch, Sylvain Gelly,
  Bernhard Sch{\"o}lkopf, and Olivier Bachem.
\newblock Challenging common assumptions in the unsupervised learning of
  disentangled representations.
\newblock In {\em Proceedings of the 36th International Conference on Machine
  Learning}, pages 4114--4124. PMLR, 2019.

\bibitem{logothetis:visual}
Nikos~K Logothetis and David~L Sheinberg.
\newblock Visual object recognition.
\newblock {\em Annual Review of Neuroscience}, 19:577--621, 1996.

\bibitem{ma:dimensionality}
Xingjun Ma, Yisen Wang, Michael~E Houle, Shuo Zhou, Sarah Erfani, Shutao Xia,
  Sudanthi Wijewickrema, and James Bailey.
\newblock Dimensionality-driven learning with noisy labels.
\newblock In {\em Proceedings of the 35th International Conference on Machine
  Learning}, pages 3355--3364. PMLR, 2018.

\bibitem{malach:decoupling}
Eran Malach and Shai Shalev-Shwartz.
\newblock Decoupling ``when to update" from ``how to update".
\newblock In {\em Advances in Neural Information Processing Systems 30}, pages
  961--971, 2017.

\bibitem{manwani:noise}
Naresh Manwani and PS~Sastry.
\newblock Noise tolerance under risk minimization.
\newblock {\em IEEE Transactions on Cybernetics}, 43(3):1146--1151, 2013.

\bibitem{natarajan:learning}
Nagarajan Natarajan, Inderjit~S Dhillon, Pradeep~K Ravikumar, and Ambuj Tewari.
\newblock Learning with noisy labels.
\newblock In {\em Advances in Neural Information Processing Systems 26}, pages
  1196--1204, 2013.

\bibitem{netzer:reading}
Yuval Netzer, Tao Wang, Adam Coates, Alessandro Bissacco, Bo~Wu, and Andrew~Y
  Ng.
\newblock Reading digits in natural images with unsupervised feature learning.
\newblock In {\em NIPS Workshop on Deep Learning and Unsupervised Feature
  Learning}, 2011.

\bibitem{patrini:forward}
Giorgio Patrini, Alessandro Rozza, Aditya Krishna~Menon, Richard Nock, and
  Lizhen Qu.
\newblock Making deep neural networks robust to label noise: A loss correction
  approach.
\newblock In {\em Proceedings of the IEEE Conference on Computer Vision and
  Pattern Recognition}, pages 1944--1952, 2017.

\bibitem{peters:elements}
Jonas Peters, Dominik Janzing, and Bernhard Schlkopf.
\newblock {\em Elements of Causal Inference: Foundations and Learning
  Algorithms}.
\newblock The MIT Press, 2017.

\bibitem{scholkopf:on}
Bernhard Sch\"{o}lkopf, Dominik Janzing, Jonas Peters, Eleni Sgouritsa, Kun
  Zhang, and Joris Mooij.
\newblock On causal and anticausal learning.
\newblock In {\em Proceedings of the 29th International Conference on Machine
  Learning}, page 459–466, 2012.

\bibitem{scholkopf:causal}
Bernhard Sch{\"o}lkopf, Dominik Janzing, Jonas Peters, Eleni Sgouritsa, Kun
  Zhang, and Joris Mooij.
\newblock On causal and anticausal learning.
\newblock {\em arXiv preprint arXiv:1206.6471}, 2012.

\bibitem{scholkopf:toward}
Bernhard Sch{\"o}lkopf, Francesco Locatello, Stefan Bauer, Nan~Rosemary Ke, Nal
  Kalchbrenner, Anirudh Goyal, and Yoshua Bengio.
\newblock Toward causal representation learning.
\newblock {\em Proceedings of the IEEE}, 109(5):612--634, 2021.

\bibitem{sohn:learning}
Kihyuk Sohn, Honglak Lee, and Xinchen Yan.
\newblock Learning structured output representation using deep conditional
  generative models.
\newblock In {\em Advances in Neural Information Processing Systems 28}, pages
  3483--3491, 2015.

\bibitem{tanaka:joint}
Daiki Tanaka, Daiki Ikami, Toshihiko Yamasaki, and Kiyoharu Aizawa.
\newblock Joint optimization framework for learning with noisy labels.
\newblock In {\em Proceedings of the IEEE Conference on Computer Vision and
  Pattern Recognition}, pages 5552--5560, 2018.

\bibitem{ullman:high}
Shimon Ullman.
\newblock High-level vision: Object recognition and visual cognition.
\newblock {\em Cell}, pages 599--606, 1996.

\bibitem{von:self}
Julius Von~K{\"u}gelgen, Yash Sharma, Luigi Gresele, Wieland Brendel, Bernhard
  Sch{\"o}lkopf, Michel Besserve, and Francesco Locatello.
\newblock Self-supervised learning with data augmentations provably isolates
  content from style.
\newblock In {\em Advances in Neural Information Processing Systems 34}, pages
  16451--16467, 2021.

\bibitem{wei:learning}
Jiaheng Wei, Zhaowei Zhu, Hao Cheng, Tongliang Liu, Gang Niu, and Yang Liu.
\newblock Learning with noisy labels revisited: A study using real-world human
  annotations.
\newblock {\em arXiv preprint arXiv:2110.12088}, 2021.

\bibitem{xia:part}
Xiaobo Xia, Tongliang Liu, Bo~Han, Nannan Wang, Mingming Gong, Haifeng Liu,
  Gang Niu, Dacheng Tao, and Masashi Sugiyama.
\newblock Part-dependent label noise: Towards instance-dependent label noise.
\newblock In {\em Advances in Neural Information Processing Systems 33}, pages
  7597--7610, 2020.

\bibitem{xia:anchor}
Xiaobo Xia, Tongliang Liu, Nannan Wang, Bo~Han, Chen Gong, Gang Niu, and
  Masashi Sugiyama.
\newblock Are anchor points really indispensable in label-noise learning?
\newblock In {\em Advances in Neural Information Processing Systems 32}, pages
  6838--6849, 2019.

\bibitem{xiao:fashion}
Han Xiao, Kashif Rasul, and Roland Vollgraf.
\newblock Fashion-mnist: a novel image dataset for benchmarking machine
  learning algorithms.
\newblock {\em arXiv preprint arXiv:1708.07747}, 2017.

\bibitem{xiao:learning}
Tong Xiao, Tian Xia, Yi~Yang, Chang Huang, and Xiaogang Wang.
\newblock Learning from massive noisy labeled data for image classification.
\newblock In {\em Proceedings of the IEEE Conference on Computer Vision and
  Pattern Recognition}, pages 2691--2699, 2015.

\bibitem{yan:learning}
Yan Yan, R{\'o}mer Rosales, Glenn Fung, Ramanathan Subramanian, and Jennifer
  Dy.
\newblock Learning from multiple annotators with varying expertise.
\newblock {\em Machine learning}, 95:291--327, 2014.

\bibitem{yang:estimating}
Shuo Yang, Erkun Yang, Bo~Han, Yang Liu, Min Xu, Gang Niu, and Tongliang Liu.
\newblock Estimating instance-dependent bayes-label transition matrix using a
  deep neural network.
\newblock In {\em Proceedings of the 39th International Conference on Machine
  Learning}, pages 25302--25312. PMLR, 2022.

\bibitem{yao:instance}
Yu~Yao, Tongliang Liu, Mingming Gong, Bo~Han, Gang Niu, and Kun Zhang.
\newblock Instance-dependent label-noise learning under a structural causal
  model.
\newblock In {\em Advances in Neural Information Processing Systems 34}, pages
  4409--4420, 2021.

\bibitem{yu:learning}
Xiyu Yu, Tongliang Liu, Mingming Gong, and Dacheng Tao.
\newblock Learning with biased complementary labels.
\newblock In {\em Proceedings of the 16th European Conference on Computer
  Vision}, pages 68--83, 2018.

\bibitem{zagoruyko:wide}
Sergey Zagoruyko and Nikos Komodakis.
\newblock Wide residual networks.
\newblock {\em arXiv preprint arXiv:1605.07146}, 2016.

\bibitem{zhang:mixup}
Hongyi Zhang, Moustapha Cisse, Yann~N Dauphin, and David Lopez-Paz.
\newblock mixup: Beyond empirical risk minimization.
\newblock {\em arXiv preprint arXiv:1710.09412}, 2017.

\bibitem{zhang:distinguishing}
Kun Zhang, Jiji Zhang, and Bernhard Sch{\"o}lkopf.
\newblock Distinguishing cause from effect based on exogeneity.
\newblock {\em arXiv preprint arXiv:1504.05651}, 2015.

\bibitem{zhang:mil}
Weijia Zhang, Xuanhui Zhang, Han-Wen Deng, and Min-Ling Zhang.
\newblock Multi-instance casual representation learning for instance label
  prediction and out-of-distribution generalization.
\newblock In {\em Advances in Neural Information Processing System 35}, 2022.

\bibitem{zhang:generalized}
Zhilu Zhang and Mert Sabuncu.
\newblock Generalized cross entropy loss for training deep neural networks with
  noisy labels.
\newblock In {\em Advances in neural information processing systems 31}, pages
  8792--8802, 2018.

\bibitem{zhou:open}
Zhi-Hua Zhou.
\newblock Open-environment machine learning.
\newblock {\em National Science Review}, 9(8):nwac123, 2022.

\end{thebibliography}

\end{document}